%% file: acl2023.tex
\pdfoutput=1

\documentclass[11pt]{article}

\usepackage[]{ACL2023}

\usepackage{times}
\usepackage{latexsym}
\usepackage{amsfonts,amssymb}

\usepackage[T1]{fontenc}

\usepackage[utf8]{inputenc}

\usepackage{microtype}

\usepackage[utf8]{inputenc}
\usepackage{booktabs}
\usepackage{graphicx}
\usepackage{CJK}
\usepackage{multirow}
\usepackage{verbatim}
\usepackage{url}
\usepackage{enumitem}
\usepackage{amsmath}
\usepackage{diagbox}
\usepackage{flushend,cuted}
\usepackage{bbm}
\usepackage{xcolor}
\usepackage{xspace}
\usepackage{algorithm}
\usepackage{algpseudocode}
\usepackage{verbatim}
\usepackage{colortbl}
\usepackage{bbding}
\usepackage{tablefootnote}

\usepackage{listings}
\usepackage{ulem}
\usepackage{makecell}
 
\newcommand{\knn}{$k$NN\xspace}
\newcommand{\knnmt}{$k$NN-MT\xspace}
\newcommand{\toolkit}{$k$NN-BOX\xspace}

\newcommand{\zwidc}[1]{\makebox[0cm][c]{#1}}

\title{$k$NN-BOX: A Unified Framework for Nearest Neighbor Generation}

\author{
    Wenhao Zhu\footnotemark[1], 
    Qianfeng Zhao\footnotemark[1], 
    Yunzhe Lv\footnotemark[1], \\
    \textbf{Shujian Huang},
    \textbf{Siheng Zhao},
    \textbf{Sizhe Liu},
    \textbf{Jiajun Chen} \\
    National Key Laboratory for Novel Software Technology, Nanjing University, China \\ 
    {\{zhuwh, qianfeng, lvyz, zhaosh, liusz\}@smail.nju.edu.cn, \{huangsj, chenjj\}@nju.edu.cn} \\
}

\begin{document}
\maketitle

\renewcommand{\thefootnote}{\fnsymbol{footnote}}
\footnotetext[1]{Equal Contribution.}
\renewcommand{\thefootnote}{\arabic{footnote}}

\input{latex/00_abstract.tex}

\input{latex/01_introduction.tex}

\input{latex/02_background.tex}

\input{latex/03_design_and_usage.tex}

\input{latex/04_experiments.tex}

\input{latex/05_conclusion.tex}

\normalem
\bibliography{anthology,custom}
\bibliographystyle{acl_natbib}

\appendix

\input{latex/06_appendix.tex}



\end{document}

%% file: latex/00_abstract.tex
\begin{abstract}
Augmenting the base neural model with a token-level symbolic datastore is a novel generation paradigm and has achieved promising results in machine translation (MT).
In this paper, we introduce a unified framework \toolkit, which enables quick development and interactive analysis for this novel paradigm.
\toolkit decomposes the datastore-augmentation approach into three modules: datastore, retriever and combiner, thus putting diverse $k$NN generation methods into a unified way.
Currently, \toolkit has provided implementation of seven popular $k$NN-MT variants, covering research from performance enhancement to efficiency optimization.
It is easy for users to reproduce these existing work or customize their own models.
Besides, users can interact with their \knn generation systems with \toolkit to better understand the underlying inference process in a visualized way.
In experiment section, we apply \toolkit for machine translation and three other seq2seq generation tasks, namely, text simplification, paraphrase generation and question generation.
Experiment results show that augmenting the base neural model with \toolkit leads to a large performance improvement in all these tasks.
The code and document of \toolkit is available at \url{https://github.com/NJUNLP/knn-box}.
\end{abstract} 

%% file: latex/01_introduction.tex
\section{Introduction}
Equipping the base neural model with a symbolic datastore is a novel paradigm for enhancing generation quality.
\citeauthor{khandelwal2021nearest} apply this paradigm in machine translation, known as \knnmt, and achieves promising results, especially in MT domain adaptation and multilingual MT.
\begin{figure}[htbp]
    \centering
    \includegraphics[scale=0.28]{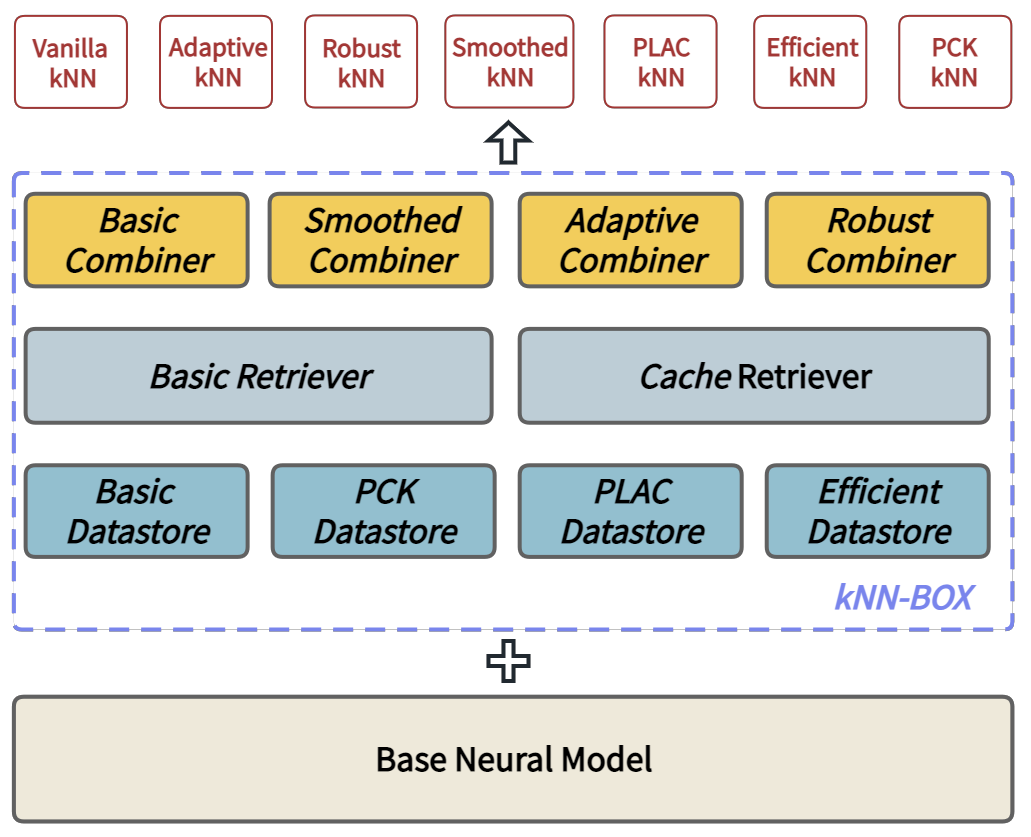}
    \caption{\toolkit decomposes the datastore-augmentation approach into three modules, namely, datastore, retriever and combiner, thus putting diverse \knn generation methods into a unified way.}
    \label{fig:uni}
\end{figure}
Afterwards, the following work keep optimizing this approach, making it a more mature methodology, e.g., dynamically deciding the usage of retrieval results~\cite{zheng2021adaptive}, building a light and explainable datastore~\cite{zhu2022knowledge}.

However, we notice that these $k$NN generation methods are implemented with diverse codebases, e.g., \textit{Fairseq}\footnote{\url{https://github.com/facebookresearch/fairseq}}, \textit{Transformers}\footnote{\url{https://github.com/huggingface/transformers}} and \textit{JoeyNMT}\footnote{\url{https://github.com/joeynmt/joeynmt}}, which hinders fair comparison between these methods and potential fusion of latest research advances.
Interpretability is another remaining issue in $k$NN generation research, as the community can still not well understand why $k$NN generation works.
More in-depth analysis needs to be conducted for this novel paradigm.

In this paper, we introduce a unified framework \toolkit for nearest neighbor generation, which supports quick development and interactive analysis.
Our framework decomposes the datastore-augmentation approach into three modules: datastore, retriever and combiner, thus putting diverse \knn generation methods into a unified way (Fig. \ref{fig:uni}).
Up till now, \toolkit has released implementation of seven popular \knnmt models, covering research from performance enhancement~\cite{khandelwal2021nearest, jiang2021learning, zheng2021adaptive, jiang2022towards} to efficiency optimization~\cite{martins2022efficient, wang2022efficient, zhu2022knowledge}, which can help users to quickly reproduce existing works and make fair comparison between them.
Moreover, users can easily fuse advanced models with \toolkit, for example, jointly using a better combiner and a lighter datastore, to achieve the best of two worlds.

Another useful feature of \toolkit is supporting visualized interactive analysis.
Via our provided web page, users can interact with their local \knn model and observe its inference process, e.g. the content and distribution of its retrieval results (Fig. \ref{fig:web}).
We hope \toolkit can help the community to better understand the interpretability problems in \knn generation, e.g., why it works.

We conduct experiments on benchmark machine translation datasets.
Experiment results show that \toolkit is a reliable platform for model reproduction and development.
In addition, we take a step further and apply \toolkit for three other seq2seq tasks, i.e., text simplification, paraphrase generation and question generation.
Experiment results show that augmenting the base neural model with \toolkit is also beneficial in these tasks, which shows the great potential of nearest neighbor generation and the wide usage of our \toolkit toolkit.

%% file: latex/02_background.tex
\section{Background: \knnmt}
Before introducing \toolkit, we recap \knnmt approach in this section.
Generally, \knnmt framework aims at memorizing translation knowledge in parallel corpus $\mathcal{C}$ into a datastore $\mathcal{D}$ and use it to augment the NMT model $\mathcal{M}$ during inference.

\noindent\paragraph{Memorizing Knowledge into Datastore} 
To extract translation knowledge, translation pair $(\mathcal{X}, \mathcal{Y})$ is fed into $\mathcal{M}$ for teacher-forcing decoding. 
At time step $t$, the continuous representation of the translation context $(\mathcal{X}, \mathcal{Y}_{<t})$, i.e. the hidden state $h_t$ from the last decoder layer, is taken as \textit{key}:
\begin{equation}
    h_t = \mathcal{M}(\mathcal{X}, \mathcal{Y}_{<t}) \nonumber
\end{equation}
and the target token $y_t$ is taken as \textit{value}.
Each \textit{key}-\textit{value} pair explicitly memorizes the translation knowledge: generating the \textit{value} token at the decoder hidden state \textit{key}. 
With a single forward pass over the entire corpus, the full datastore $\mathcal{D}$ can be constructed:
\begin{equation}
\label{eq:datastore}
    \mathcal{D} = \{(h_t, y_t) ~|~ \forall y_t \in \mathcal{Y}, (\mathcal{X}, \mathcal{Y})\in \mathcal{C}\},
\end{equation}

\noindent\paragraph{Generating with Memorized Knowledge}
The constructed datastore is then combined with the base NMT model as an augmentation memory.
During inference, the NMT model retrieves related knowledge from the datastore to adjust its own translation prediction. 

Specifically, the NMT model uses the contextualized representation of the test translation context $(\mathcal{X}, \mathcal{Y}_{<t})$ to query the datastore for nearest neighbor representations and the corresponding target tokens $\mathcal{N}_k=\{(h^j, y^j)\}_{j=1}^k$.
The retrieved entries are then converted to a distribution over the vocabulary:
\begin{equation}
\label{eq:knn}
p_{\text{knn}}(y|\mathcal{X}, \mathcal{Y}_{<t}) \propto \sum_{(h^j, y^j)\in \mathcal{N}_k} \mathbbm{1}(y=y^j)\cdot s(h_t, h^j)
\end{equation}
where $s$ measures the similarity between $h_t$ and $h^j$:
\begin{equation}
s(h_t, h^j) = \exp[{\frac{-d(h_t,h^j)}{T}}] \nonumber
\end{equation}
Here, $d$ denotes $L_2$-square distance and $T$ is the temperature.
In the end, the output distribution of the NMT model and symbolic datastore are interpolated with the weight $\lambda$:
\begin{equation}
\label{eq:combine}
\begin{split}
p(y|\mathcal{X}, \mathcal{Y}_{<t}) &= \lambda\cdot \ p_{\textrm{knn}}(y|\mathcal{X}, \mathcal{Y}_{<t}) \\
& + (1-\lambda)\cdot \ p_{\textrm{nmt}} (y|\mathcal{X}, \mathcal{Y}_{<t})
\end{split}
\end{equation}

\noindent\paragraph{Recent Advances in \knnmt} 
To make \knnmt more effective, efficient and explainable, various methods have been devised.
\citet{zheng2021adaptive} and \citet{jiang2022towards} propose to dynamically decide the usage of retrieval results to exclude potential noise in nearest neighbors.
\citet{jiang2021learning} explore the setting of multi-domain adaptation and remedy the catastrophic forgetting problem.
Inspired by \citet{he2021efficient}, \citet{martins2022efficient} introduce three ways to improve the efficiency of $k$NN-MT, i.e. dimension reduction, datastore pruning and adaptive retrieval.
Later, \citet{wang2022efficient} propose to reduce dimension and prune datastore with a learnable network.
Recently, \citet{zhu2022knowledge} explore the interpretability issue in \knnmt and builds a light and more explainable datastore according to the capability of the NMT model.

%% file: latex/03_design_and_usage.tex
\section{Unified Framework: \toolkit}
This sections describes how we design and implement \toolkit, and introduce how users run \toolkit for developing \knn generation models and interacting with the deployed model visually.

\subsection{Design and Implementation}
\begin{figure}[htbp]
    \centering
    \includegraphics[width=0.50\textwidth]{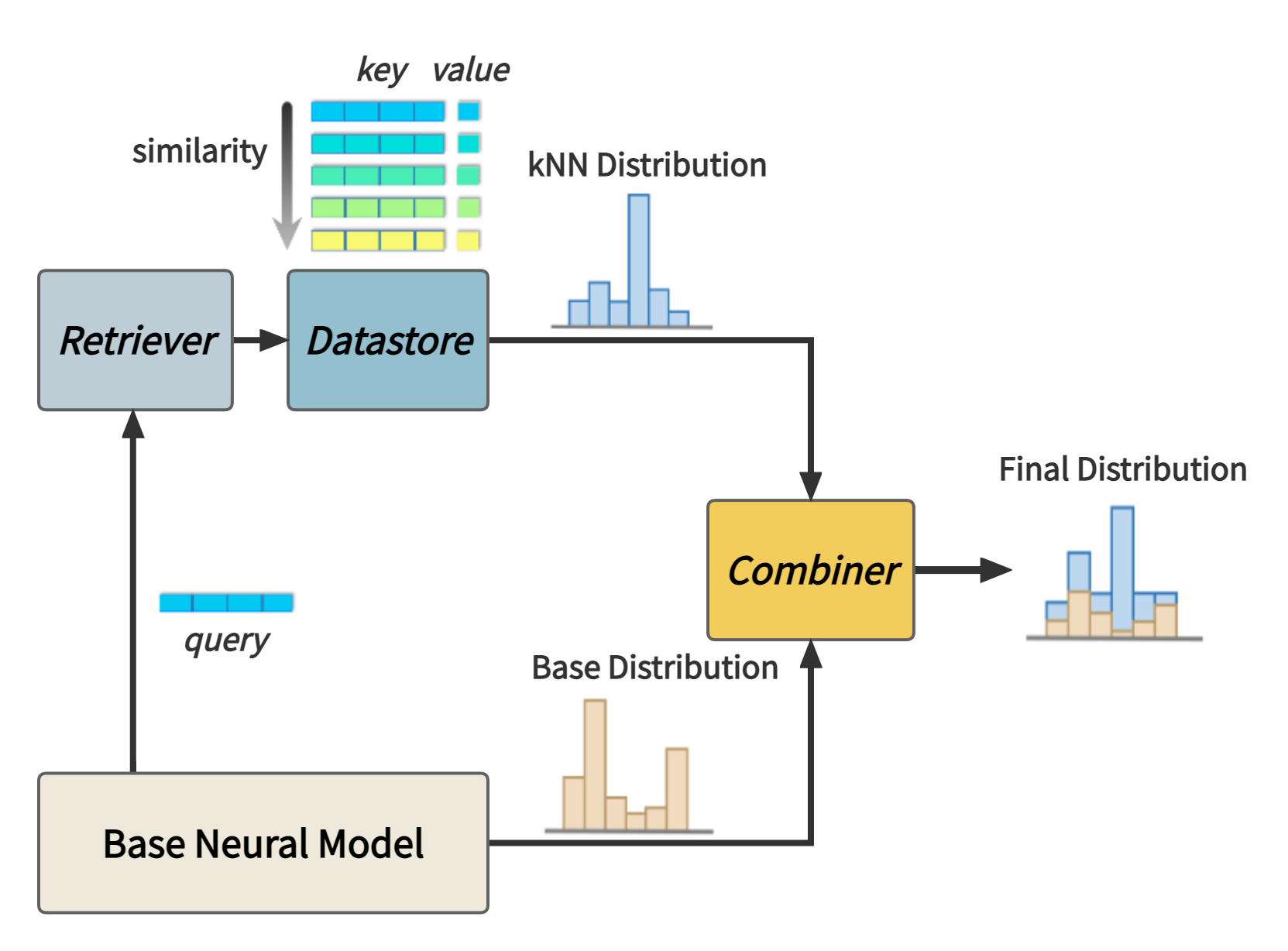}
    \caption{Overall workflow of augmenting the base neural model with \toolkit .}
    \label{fig:flow}
\end{figure}

We develop \toolkit based on the most widely-used generation framework \textit{Fairseq}, thus making it easy to apply \toolkit for other generation tasks.
The overall workflow of \toolkit is illustrated in Figure \ref{fig:flow}.
For better compatibility and extensibility, we decompose the datastore-augmentation approach into three modules: \texttt{Datastore}, \texttt{Retriever} and \texttt{Combiner}, where each module has its own function:
\begin{itemize}[itemsep=1pt]
    \item \texttt{Datastore}: save generation knowledge as \textit{key}-\textit{values} pairs (Eq. \ref{eq:datastore}).
    \item \texttt{Retriever}: retrieve nearest neighbors from the datastore during inference.
    \item \texttt{Combiner}: convert retrieval results to a distribution (Eq. \ref{eq:knn}) and interpolate the output distribution of the NMT model and symbolic datastore (Eq. \ref{eq:combine}).
\end{itemize}

This design enables diverse \knn models to be implemented in a unified way.
For a specific \knn variant, it usually makes a modification on one of the three modules, compared to vanilla \knn generation model.
Therefore, users can customize the corresponding module and quickly develop the desired \knn model.

Supporting visual interactive analysis is another useful feature of \knnmt.
By saving intermediate computation results, we enable \toolkit to visualize the inference process.
We hope this feature will help users to better understand their own \knnmt model's strengths and weaknesses, instead of using it as a black box.

\subsection{Usage}

\noindent\paragraph{Reproducing Existing Work} 
Until now, \toolkit has released implementation of seven popular \knnmt models \footnote{They are vanilla \knnmt~\cite{khandelwal2021nearest}, Adaptive \knnmt~\cite{zheng2021adaptive}, Smoothed \knnmt~\cite{jiang2021learning}, Robust \knnmt~\cite{jiang2022towards}, PCK \knnmt~\cite{wang2022efficient}, Efficient \knnmt~\cite{martins2022efficient}, PLAC \knnmt~\cite{zhu2022knowledge}.}, covering research from performance enhancement to efficiency optimization.
Besides, \toolkit has also provided the corresponding shell scripts to run them, enabling users to quickly reproduce existing work.
Detailed guidance can be found in \texttt{README.md}.

\noindent\paragraph{Developing New Models} 
\toolkit is designed not only for reproduce existing work, but also for developing new models on new tasks.
For each module, users can pick one of its implementation from \toolkit or customize their own version, and combine three modules together to build a new \knn model. 
In this process, only few lines of codes needs to be added, which will save users a lot of time.
More importantly, this implementation fashion enables users to easily build a fused model, e.g., combining the most explainable datastore (\texttt{PLACDatstore}) with the strongest combiner (\texttt{RobustCombiner}).
To perform generation tasks other than machine translation, users only need to switch the training corpus to build a task-specific datastore.

\begin{figure*}[ht]
    \centering
    \includegraphics[width=0.8\textwidth]{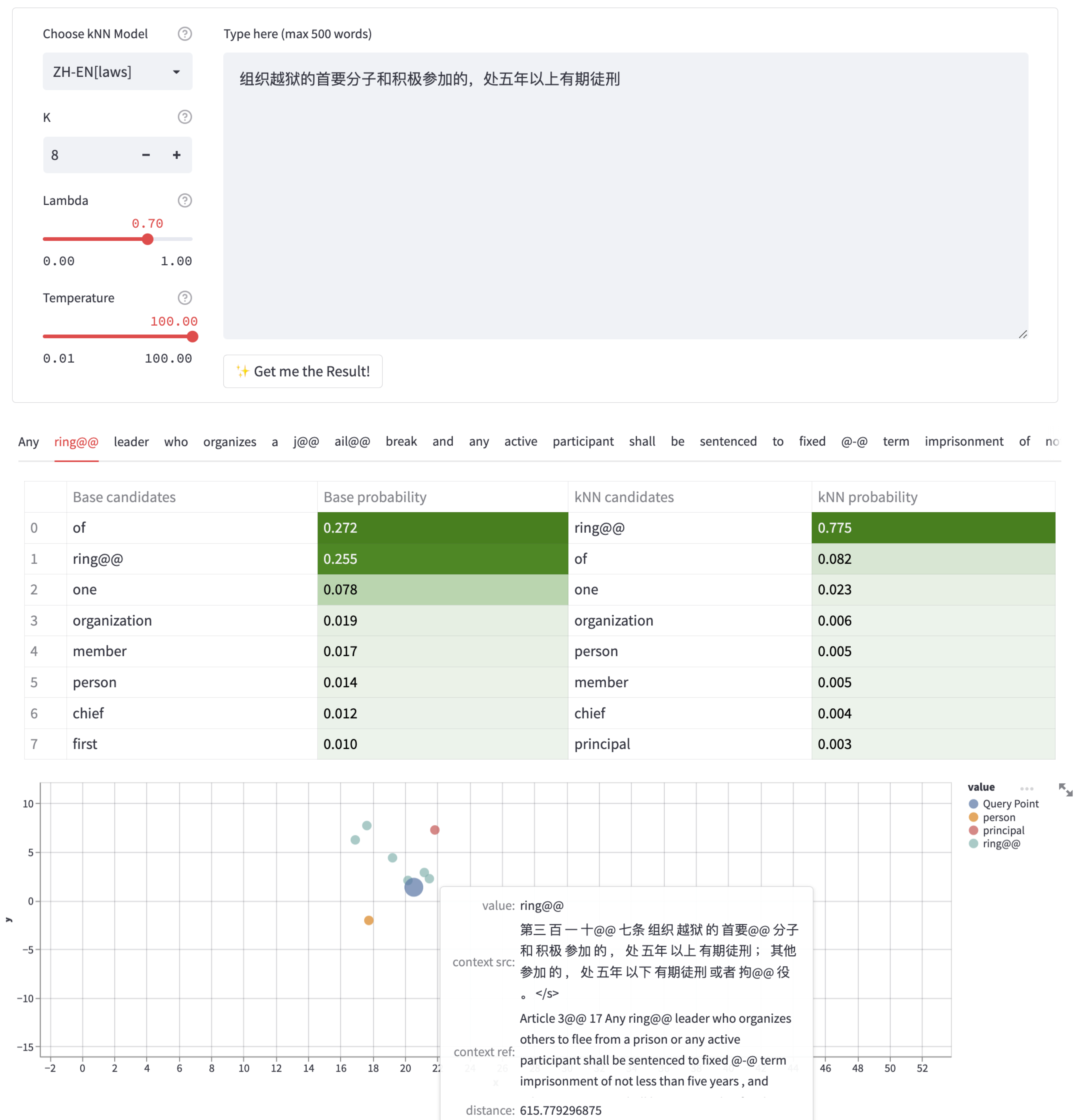}
    \caption{A screenshot of interactive interface provided by \toolkit, where users can interact with their own $k$NN model and analyze its inference process visually. The upper panel allows users to type in text and tune hyperparameters. The middle panel displays the generation result and prediction distribution of each decoding step. The bottom panel shows the relative distribution of query and retrieval results, and more detailed information of each nearest neighbor.}
    \label{fig:web}
\end{figure*}

\noindent\paragraph{Visualizing Translation Process} 
By running our provided script to launch a web page (shown in Fig. \ref{fig:web}), users can interact with their \knn model visually.
In the upper input window, user can type in text and tune generation hyperparameters in the upper-left panel.
Then the generation results will be displayed.
Taking \knnmt as an example, after clicking a word in the translation, users can see the translation probability given by both NMT model and \knnmt model.
Moreover, detailed information of the retrieved datastore entries will be displayed in the bottom panel.
By selecting on a certain nearest neighbor point, users can see the corresponding value token, translation context and \textit{query}-\textit{key} distance.
Overall, the visualization page can help user to interact with their \knn generation model and explore its inner working process.

%% file: latex/04_experiments.tex
\section{Experiments}
To evaluate the effectiveness of \toolkit, we conduct experiments on machine translation and three other seq2seq tasks.
We introduce experimental settings in Section \ref{sec:setting} and reports experiment results in Section \ref{sec:results}.

\subsection{Experimental Settings}
\label{sec:setting}
\noindent\paragraph{Dataset}
For machine translation, we adopt four German-English OPUS datasets \footnote{\url{https://opus.nlpl.eu/}} (Medical, Law, IT and Koran) \cite{tiedemann2012parallel}, which are used in almost all \knnmt work.
We use TED dataset \footnote{\url{https://github.com/neulab/word-embeddings-for-nmt}} \cite{qi2018when} to evaluate \toolkit on multilingual machine translation \footnote{We evaluate English-centric translation performance on ten languages: Cs, Da, De, Es, Fr, It, Nl, Pl, Pt, Sv}.
Moreover, we conduct experiments on two text simplification dataset: Newsela-Auto \footnote{\url{https://newsela.com/data/}} and Wiki-Auto \footnote{\url{https://github.com/chaojiang06/wiki-auto/tree/master/wiki-auto/ACL2020/}} \cite{jiang-etal-2020-neural}, a paraphrase generation dataset: QQP \footnote{\url{https://www.kaggle.com/c/quora-question-pairs}}, and a question generation dataset: Quasar-T \footnote{\url{https://github.com/bdhingra/quasar}} \cite{dhingra2017quasar} to demonstrate effectiveness of \toolkit on these generation tasks.

\begingroup
\renewcommand{\arraystretch}{1.2} 
\begin{table*}[ht]
    \footnotesize
    \centering
    \scalebox{0.9}{
    \begin{tabular}{p{3cm}<{\centering}|p{3.5cm}<{\centering}|p{0.8cm}<{\centering}p{0.8cm}<{\centering}|p{0.8cm}<{\centering}p{0.8cm}<{\centering}|p{0.8cm}<{\centering}p{0.8cm}<{\centering}|p{0.8cm}<{\centering}p{0.8cm}<{\centering}}
    \toprule
    \multirow{2}{*}{\zwidc{Model}} & \multirow{2}{*}{\zwidc{Reference}} & \multicolumn{2}{c|}{\zwidc{\textbf{Law}}} & \multicolumn{2}{c|}{\zwidc{\textbf{Medical}}}    & \multicolumn{2}{c|}{\zwidc{\textbf{IT}}} & \multicolumn{2}{c}{\zwidc{\textbf{Koran}}} \\
    & & \zwidc{Scale$\downarrow$} & \zwidc{BLEU$\uparrow$} & \zwidc{Scale$\downarrow$} & \zwidc{BLEU$\uparrow$} & \zwidc{Scale$\downarrow$} & \zwidc{BLEU$\uparrow$} & \zwidc{Scale$\downarrow$} & \zwidc{BLEU$\uparrow$} \\
    \midrule
    Base Neural Model & ~\citealp{ng2019facebook}             & -     & 45.5 & 100\% & 40.0 & -     & 38.4 & -     & 16.3 \\
    \midrule
    Vanilla \knnmt    & ~\citealp{khandelwal2021nearest}      & 100\% & 61.3 & 100\% & 54.1 & 100\% & 45.6 & 100\% & 20.4 \\
    Adaptive \knnmt   & ~\citealp{zheng2021adaptive}          & 100\% & 62.9 & 100\% & 56.1 & 100\% & 47.2 & 100\% & 20.3 \\
    Smoothed \knnmt   & ~\citealp{jiang2021learning}          & 100\% & 63.3 & 100\% & 56.8 & 100\% & 47.7 & 100\% & 19.9 \\
    Robust \knnmt     & ~\citealp{jiang2022towards}           & 100\% & 63.6 & 100\% & 57.1 & 100\% & 48.6 & 100\% & 20.5 \\
    PCK \knnmt        & ~\citealp{wang2022efficient}          &  90\% & 62.8 &  90\% & 56.4 &  90\% & 47.4 &  90\% & 19.4 \\
    Efficient \knnmt  & ~\citealp{martins2022efficient}       & 57\%  & 59.9 & 58\%  & 52.3 & 63\%  & 44.9 & 66\%  & 19.9 \\
    PLAC \knnmt       & ~\citealp{zhu2022knowledge}           & 55\%  & 62.8 & 55\%  & 56.2 & 60\%  & 47.0 & 75\%  & 19.9 \\
    \bottomrule
    \end{tabular}
    }
    \caption{Some works implemented by \toolkit. Scale refers to the relative datastore size compared to a full datastore that covers all target language token occurrences in the parallel corpus. Smaller scale means a lighter datastore and higher BLEU means better translation quality.}
    \label{tab:performance}
\end{table*}
\endgroup

\begingroup
\renewcommand{\arraystretch}{1.2} 
\begin{table*}[ht] 
    \footnotesize 
    \centering 
    \scalebox{1.0}{ 
    \begin{tabular}{lc|ccccccccccc} 
    \toprule 
    Directions & Model & \zwidc{\textbf{Avg.}}	& \textbf{Cs}	& \textbf{Da}	& \textbf{De}	& \textbf{Es}	& \textbf{Fr}	& \textbf{It}	& \textbf{Nl}	& \textbf{Pl}	& \textbf{Pt}	& \textbf{Sv}\\ 
    \midrule 
    \multirow{2}{*}{$\textbf{En}\to\textbf{X}$}  & M2M-100 	& 29.1 	& 20.7 	& 36.2 	& 26.7 	& 35.1 	& 33.7 	& 29.8 	& 27.7 	& 15.6 	& 31.9 	& 33.7 \\ 
                                                 & + \toolkit & 32.6  & 22.3  & 40.2  & 29.5  & 39.2  & 38.7  & 33.5  & 31.9  & 17.9  & 37.1  & 36.0 \\
    \midrule \multirow{2}{*}{$\textbf{X}\to\textbf{En}$}  & M2M-100 & 33.4 	& 27.5 	& 40.0 	& 31.8 	& 36.6 	& 35.1 	& 33.4 	& 31.9 	& 21.1 	& 38.9 	& 37.3 \\
                                                          & + \toolkit& 37.7  & 31.3  & 44.5  & 37.1  & 42.0  & 40.4  & 38.4  & 36.2  & 24.9  & 41.8  & 41.0  \\
    \bottomrule
    \end{tabular} 
    } 
    \caption{Effect of augmenting M2M100 with \toolkit on multilingual TED dataset. For brevity, we only show the effect of applying Robust $k$NN with \toolkit. ``$\text{En}\to\text{X}$'' and ``$\text{X}\to\text{En}$'' denotes translating English into other languages and translating other languages into English respectively.} 
    \label{tab:multilingual}
\end{table*}
\endgroup

\noindent\paragraph{Base Neural Model}
On OPUS dataset, we follow previous \knnmt work and use the winner model of 
WMT’19 De-En news translation task \cite{ng2019facebook} as the base model.
On multilingual TED dataset, we use M2M100 \cite{fan2021m2m} as the base model, which is a many-to-many multilingual translation model.
On the rest of dataset, Transformer \cite{vaswani2017attention} is used as the base model. 

\begingroup
\renewcommand{\arraystretch}{1.2} 
\begin{table*}[ht]
    \footnotesize
    \centering
    \scalebox{1.0}{
    \begin{tabular}{p{4.0cm}<{\centering}|p{2.5cm}<{\centering}|p{1.5cm}<{\centering}|p{1.5cm}<{\centering}p{1.5cm}<{\centering}}
    \toprule
    \zwidc{Task}    & \zwidc{Dataset}   & \zwidc{Metric}    & \zwidc{Base Model}    & \zwidc{\toolkit} \\
    \midrule
    \multirow{2}{*}{\makecell{Text Simplification}}
                    & Wiki-Auto    & SARI & 38.6 & \textbf{39.4} \\
                    & Newsela-Auto & SARI & 35.8 & \textbf{38.2} \\
    \midrule
    \multirow{1}{*}{\makecell{Paraphrase Generation}}
                    & QQP                  & BLEU      & 28.4      & \textbf{29.5}  \\
    \midrule
    \multirow{1}{*}{\makecell{Question Generation}}
                    & Quasar-T             & BLEU      & 9.6       & \textbf{15.7}  \\
    \bottomrule
    \end{tabular}
    }
    \caption{The performance of applying \toolkit on three other seq2seq tasks: text simplification, paraphrase generation and question generation. Here, we apply the vanilla \knn generation method for augmentation. Bold text indicates the higher score across two models. Augmenting base neural models in these tasks with \toolkit also bring large performance improvement.}
    \label{tab:tasks}
\end{table*}
\endgroup
\begingroup
\renewcommand{\arraystretch}{1.2} 
\begin{table*}[ht]
    \footnotesize
    \centering
    \scalebox{1.0}{
    \begin{tabular}{p{3cm}<{\centering}p{3cm}<{\centering}p{3cm}<{\centering}|p{1cm}<{\centering}p{1cm}<{\centering}}
    \toprule
    \zwidc{Datastore}   & \zwidc{Retriever}  & \zwidc{Combiner} & \zwidc{Scale$\downarrow$} & \zwidc{BLEU$\uparrow$} \\
    \midrule
    \texttt{BasicDatastore}               & \texttt{BasicRetriever} & \texttt{BasicCombiner}     & 100\%  & 61.3 \\
    \texttt{PCKDatastore}                 & \texttt{BasicRetriever} & \texttt{AdaptiveCombiner}  &  90\%  & 62.8 \\
    \texttt{EfficientDatastore}           & \texttt{BasicRetriever} & \texttt{AdaptiveCombiner}  &  57\%  & 61.5 \\
    \texttt{EfficientDatastore}           & \texttt{BasicRetriever} & \texttt{RobustCombiner}    &  57\%  & 61.8 \\
    \texttt{PLACDatastore}                & \texttt{BasicRetriever} & \texttt{AdaptiveCombiner}  &  55\%  & 62.8 \\
    \texttt{PLACDatastore}                & \texttt{BasicRetriever} & \texttt{RobustCombiner}    &  55\%  & 63.7 \\
    \bottomrule
    \end{tabular}
    }
    \caption{Effect of fusing advanced datastore and combiner. Smaller scale means a lighter datastore and higher BLEU means better translation quality.}
    \label{tab:combination}
\end{table*}
\endgroup

\noindent\paragraph{Metric}
We use BLEU score calculated by \textit{sacrebleu}\footnote{\url{https://github.com/mjpost/sacrebleu}} to evaluate the generation quality for all tasks except text simplification, where we use SARI score \cite{xu-etal-2016-optimizing} calculated by \textit{easse}\footnote{\url{https://github.com/feralvam/easse}} to evaluate simplification quality.

\subsection{Main Results}
\label{sec:results}

\noindent\paragraph{\toolkit can help user to quickly augment the base NMT model with $k$NN methods.}
By running our provided shell scripts, users can quickly reproduce existing \knnmt models.
Table \ref{tab:performance} show the translation performance of these models on OPUS dataset.
We see that augmenting the base neural machine translation model with a datastore brings significant performance enhancement.
Among these methods, Robust \knnmt achieves the highest BLEU scores, and PLAC \knnmt builds a lightest datastore while maintaining translation performance.
Table \ref{tab:multilingual} reports experiment results on TED dataset.
We can see that applying \toolkit brings large performance improvement on all translation directions. 

Besides, we also carefully compare the reproduced results with the results produced by the original implementation.
We find that two groups of results are well-aligned (shown in Appendix \ref{sec:align}), demonstrating that \toolkit is reliable platform for reproducing \knnmt models.

\noindent\paragraph{\toolkit shows great potential in other seq2seq generation tasks as well}
Apart from machine translation task, we further evaluate \toolkit on three other seq2seq tasks: text simplification, paraphrase generation and question generation.
Experiment results are shown in Table \ref{tab:tasks}.
Augmenting the base neural model with \toolkit brings performance enhancement in all three tasks.
The performance improvement on three tasks is up to 2.4 SARI, 1.1 BLEU and 6.1 BLEU respectively, which shows the great potential of studying datastore-augmentation in generation tasks and the wide usage of our toolkit.

\noindent\paragraph{\toolkit accelerates the fusion of lasted research advances}
A potential drawback of implementing \knnmt with diverse codebases is hindering the fusion of lasted research advances.
With \toolkit, research advances on \texttt{Datastore}, \texttt{Combiner} and \texttt{Retriever} can be fused conveniently.
Table \ref{tab:combination} shows the performance of partial mixed models on OPUS-Law dataset, where we jointly use different datastore and combiner.
We can see that using \texttt{PLACDatastore} and \texttt{RobustCombiner} together achieve strong translation performance with a much smaller datastore.

%% file: latex/05_conclusion.tex
\section{Conclusion and Future Work}
This paper introduces \toolkit, an open-sourced toolkit for nearest neighbor generation.
\toolkit decomposes datastore-augmented approach into three decoupled modules: \texttt{Datastore}, \texttt{Retriever} and \texttt{Combiner}, thus putting diverse \knn generation methods into a unified way.
\toolkit provides implementation of several \knnmt models, covering research from performance enhancement and efficiency optimization, which can help users to quickly reproduce existing work.
\toolkit also enjoys great extensibility, which can be used to develop new models and be applied for new generation tasks.
More importantly, \toolkit supports users to interact with their deployed model in a visualized way, which enables in-depth analysis on the inner working process of the model.
In experiment section, we show that \toolkit can not only be applied for enhancing neural machine translation model, but also for enhancing neural generation model in other seq2seq tasks, like text simplification, paraphrase generation and question generation.

In the future, we will keep update this toolkit to provide implementation of more retrieve-and-generate methods and optimize the framework to make it more user-friendly, and explore the possibility to apply \toolkit for long-range sequence generation, e.g., document-level machine translation.

%% file: latex/06_appendix.tex
\section{Performance Alignment between \toolkit's implementation and original implementation}
\label{sec:align}
Table \ref{tab:performance_alignment} compares the reproduced results with \toolkit and the results produced by the original implementation, where the same base neural model and the same dataset is used.
Comparison results show that there is only a minor gap between two groups of results, demonstrating that the reliability of \toolkit.
\begin{table}[htp]
    \footnotesize 
    \centering
    \scalebox{1.0}{
    \begin{tabular}{p{3cm}|p{0.6cm}<{\centering}p{0.6cm}<{\centering}p{0.6cm}<{\centering}p{0.6cm}<{\centering}}
    \toprule
    \hspace{1cm}Model                               & \zwidc{Law} & \zwidc{Medical}    & \zwidc{IT} & \zwidc{Koran} \\
    \midrule
    Base NMT\tablefootnote{\url{https://github.com/facebookresearch/fairseq}}
    & 45.5	& 40.0 & 38.4	& 16.3	\\
    $\hookrightarrow$ \toolkit	        & 45.5	& 40.0 & 38.4	& 16.3	\\
    \midrule
    Vanilla \knnmt\tablefootnote{\url{https://github.com/urvashik/knnmt}}
    & 61.3	& 54.1 & 45.6	& 20.4	\\
    $\hookrightarrow$ \toolkit	        & 61.3	& 54.1 & 45.6	& 20.4	\\
    \midrule
    Adaptive \knnmt\tablefootnote{\url{https://github.com/zhengxxn/adaptive-knn-mt}\label{adaptve}}
    & 62.9	& 56.6 & 47.6	& 20.6	\\
    $\hookrightarrow$ \toolkit	        & 62.9	& 56.1 & 47.2	& 20.3	\\
    \midrule
    PCK \knnmt \tablefootnote{\url{https://github.com/tjunlp-lab/PCKMT}}
    & 63.1	& 56.5 & 47.9	& 19.7	\\
    $\hookrightarrow$ \toolkit          & 62.8	& 56.4 & 47.4	& 19.4	\\
    \midrule
    Robust \knnmt\tablefootnote{\url{https://github.com/DeepLearnXMU/Robust-knn-mt}}
    & 63.8	& 57.0 & 48.7	& 20.8  \\
    $\hookrightarrow$ \toolkit          & 63.6	& 57.1 & 48.6	& 20.5	\\
    \bottomrule
    \end{tabular}
    }
    \caption{BLEU scores of original implementation and \toolkit's implementation. ``$\hookrightarrow$ \toolkit' denotes the reults reproduced using our framework.}
    \label{tab:performance_alignment}
\end{table}